\documentclass[11pt,a4paper]{article}
\usepackage[hyperref]{acl2021}
\usepackage{times}
\usepackage{latexsym}
\usepackage{graphicx}

\usepackage{microtype}
\usepackage{amsmath}
\usepackage{tabularx}
\usepackage{caption}
\usepackage{subcaption}
\usepackage{enumitem}
\usepackage{array}
\usepackage{makecell}
\usepackage{hyperref}

\newcommand\sysname{\textsc{TweeNLP}}
\aclfinalcopy % Uncomment this line for the final submission

\title{TweeNLP: A Twitter Exploration Portal for Natural Language Processing}

\author{Viraj Shah, Shruti Singh \and Mayank Singh \\ Indian Institute of Technology Gandhinagar, Gujarat, India \\ \{shah.viraj, singh\_shruti, singh.mayank\}@iitgn.ac.in}

\date{}

\begin{document}
\maketitle
\begin{abstract}
We present \sysname{}, a one-stop portal that organizes Twitter's natural language processing (NLP) data and builds a visualization and exploration platform. It curates 19,395 tweets (as of April 2021) from various NLP conferences and general NLP discussions. It supports multiple features such as TweetExplorer to explore tweets by topics, visualize insights from Twitter activity throughout the organization cycle of conferences, discover popular research papers and researchers. It also builds a timeline of conference and workshop submission deadlines. We envision \sysname{} to function as a collective memory unit for the NLP community by integrating the tweets pertaining to research papers with the NLPExplorer scientific literature search engine. The current system is hosted at \href{http://nlpexplorer.org/twitter/CFP}{URL}.
\end{abstract}

\section{Introduction}
\begin{figure*}
    [!ht]
    \centering
    \includegraphics[width=1.5\columnwidth]{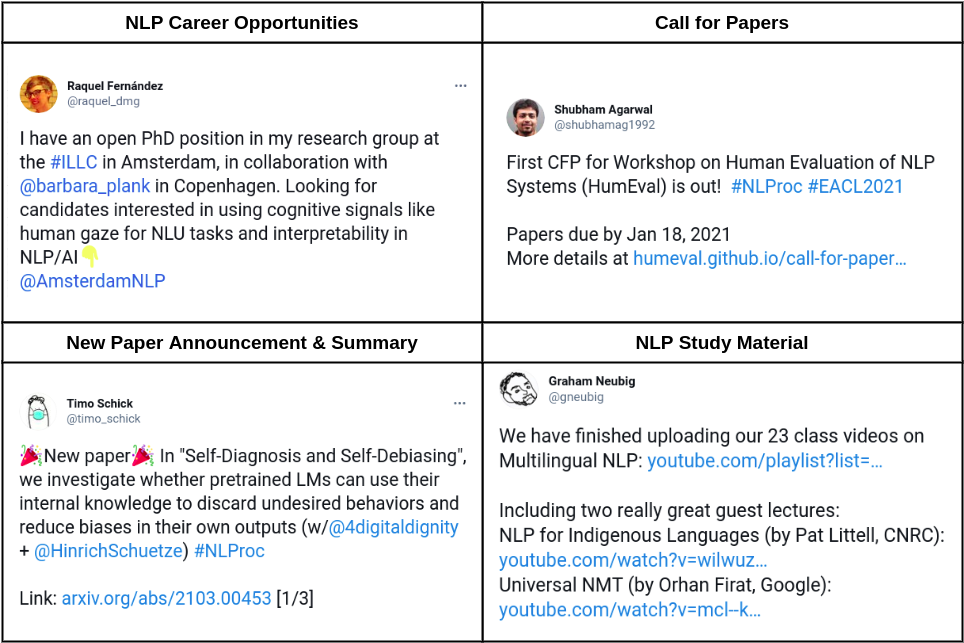}
    \caption{A sample of diverse natural language processing tweets.}
    \label{fig:tweet-example}
\end{figure*}

Online communication channels have become popular in the Internet era, and several online communities of like-minded people have evolved around these channels. 
% Users have multiple alternatives for communication channels for different purposes. 
For example, communities such as  Stack Overflow and AskUbuntu are question-answering forums; Twitter and Reddit are content-sharing forums. These forums over the years have provided a platform for novice users to learn from the experts, facilitated discussions among the community members, and have over the years accumulated a rich database of questions, answers, and discussions.

According to the theory of diffusion of innovation proposed by \citet{UBHD2028615}, the communication channel is one of the four main elements influencing the spread of a new idea. Notably, the communication channel serves as a collective long-term memory or a knowledge archive of the community, which any member can access to study the community's stance on diverse topics at any point in time.

Although several mailing lists, slack channels, and subreddits exist for communication, most natural language processing (henceforth NLP) community discussions are primarily carried out on Twitter due to its open accessibility and wider reach. 
Announcements of calls for papers and submission deadlines, recently accepted papers, interesting talks and seminars, lecture videos, and tutorials on various topics are often posted on Twitter. These are a great medium to stay updated on the recent developments in the NLP field. 
It is also a medium for researchers to engage in informal research discussions which might be unreported in official publications. 
We present a sample of diverse NLP tweets in Figure~\ref{fig:tweet-example} to emphasise the utility of the platform.

However, unlike subreddits or communities like Stack Overflow and AskUbuntu, Twitter is not an exclusive channel for NLP discussions. Exclusive channels provide users a one-stop destination for their interests and allow extremely topic-specific exploration. While Twitter allows search by hashtags to narrow down to specific topics, the usage of hashtags is highly irregular. Furthermore, Twitter is more suited to live discussions and less suitable for maintaining a snapshot of the discussions taking place in the online community. 
Relevant Twitter discussions about specific research papers are often forgotten in the long run because there is no infrastructure to link these discussions with the papers on the proceedings archives or research paper search engines.
In an attempt to address these issues, we extend the functionality of NLPExplorer~\cite{parmar2020nlpexplorer} platform by integrating~\sysname{}~with it. NLPExplorer is a portal for searching, and visualizing NLP research volume based on the ACL Anthology~\cite{ACLAnth}. 
In our current work, we build an automatic pipeline for curating NLP tweets and build a one-stop portal - \emph{\sysname{}}, for the search and browsing of NLP discussion on Twitter. The system has curated 19,395 NLP tweets as of April 2021.
% We also link specific tweets to the original papers page so as to preserve the rich twitter discussions about papers. 
% We posit that~\sysname~will serve as a collective long-term memory for the NLP community, which any member can access at any point in time to study the discussions by the community on various diverse topics. 

\sysname{} organizes NLP tweets into topics: (i) New paper announcements, (ii) Call for Paper announcements, (iii) Reading Materials \& Tutorials, (iv) Career Opportunities, (v) Talks \& Seminars, and (vi) Others. Topic-wise tweets are presented via dashboards for easy exploration. \sysname{}~supports dashboards to browse through popular NLP tweets in the previous week and the month. We construct a CFP Timeline from ‘Call for Papers’ announcements on Twitter and arrange it according to the upcoming submission deadlines of various workshops and conferences. We link the research paper tweets to the research paper's metadata, accessible via the NLPExplorer paper discovery feature. We also build live Conference Visualization dashboards, which curate tweets about the conference schedule, ongoing talks, poster sessions, and interesting papers at the conference, and present statistics such as popular hashtags, users, tweet languages, etc.
% It also presents statistics such as popular hashtags, users, tweet languages, and the distribution of conference tweets over time.

We integrate \sysname{} with NLPExplorer (Section~\ref{sec:nlpexp}) to build a joint-portal that aims to bridge the gap between published research and its informal communication on the social media platform Twitter. Our automatic data curation pipeline and the architecture of the system is described in Section~\ref{sec:data} and Section~\ref{sec:architecture} respectively. We describe the features of~\sysname~in detail in Section~\ref{sec:features}. In Section~\ref{sec:relwork}, we discuss previous works in organizing the NLP literature and visualization of research papers.

\section{NLPExplorer}
\label{sec:nlpexp}
NLPExplorer\footnote{http://nlpexplorer.org/}~\cite{parmar2020nlpexplorer} is an automatic portal for indexing, searching, and visualizing Natural Language Processing research volume. It presents multiple paper, venue, and author statistics, including paper citation distribution, paper topic distribution, authors, their field of study, their citation distributions, etc. It also presents category information of research papers into various topics broadly arranged in five categories: (i) Linguistic Target (Syntax, Discourse, etc.), (ii) Tasks (Tagging, Summarization, etc.), (iii) Approaches (unsupervised, supervised, etc.), (iv) Languages (English, Chinese, etc.) and (v) Dataset types (news, clinical notes, etc.). The current snapshot consists of 75k research papers and 50k authors. Since its inception, it has been accessed by more than 7.3k users having a close to 9.7k sessions.

\begin{figure}[!t]
    \centering
    \includegraphics[width=0.9\columnwidth]{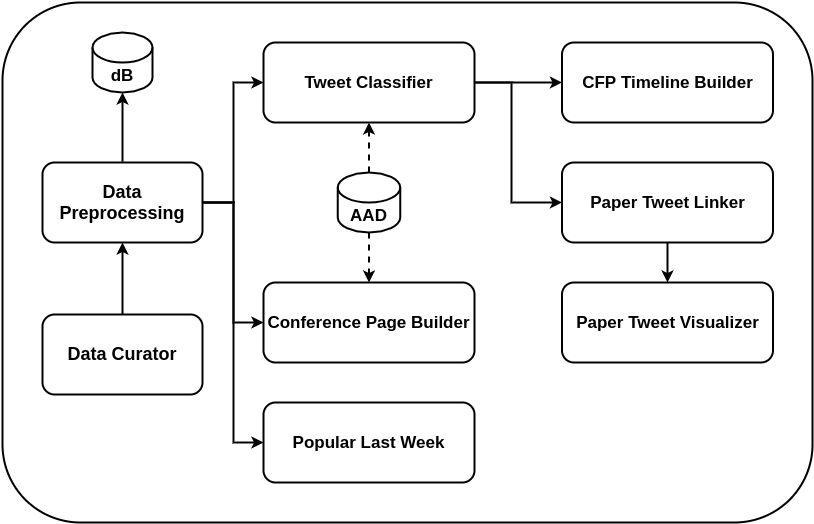}
    \caption{The architecture of \sysname{}. Arrow directions denote the flow of data. AAD represents the ACL Anthology Dataset which is the other data source apart from Twitter.}
    \label{fig:pipeline}
\end{figure}

\section{Dataset}
\label{sec:data}
We curate the dataset from two primary sources:
\subsection{Twitter}
We curate the Twitter data using the open-source library \emph{Twint\footnote{\url{https://github.com/twintproject/twint}}} by retrieving tweets with the hashtag NLProc. We also curate tweets with NLP conference hashtags such as \#acl2020, \#emnlp2020, etc. The list of NLP conferences is compiled via ACL Anthology. Our system is scheduled to download the Twitter data for each day automatically. For ongoing conferences, our system curates new tweets every hour to continually update the \emph{Conference Visualizer} page. The current snapshot (as of April 2021) contains data since October 2017 (around 1300 days) and consists of 19,395 tweets.

\subsection{ACL Anthology}
We curate the conference and journal names and URLs from the ACL Anthology github repository\footnote{\url{https://github.com/acl-org/acl-anthology}}. We also curate the paper titles and their links. Tweets are collected periodically every day, and the system checks for paper mentions in the tweets by substring matching the paper URLs collected from the ACL Anthology github repository.

\section{Architecture}
\label{sec:architecture}
We present the pipeline of our system in Figure~\ref{fig:pipeline}. The Data Curator module curates tweets daily. The curated tweets are processed before we perform further steps. The following modules process tweets: (i) Tweet Classifier, (ii) Conference Page Builder, (iii) CFP Timeline Builder, and (iv) Paper Tweet Linker. We describe the tweet processing modules in detail below:
\begin{enumerate}[noitemsep,nolistsep]
    \item \emph{Tweet Classifier}: The Tweet Classifier module classifies a tweet into one of the six topics: (i) New Paper Announcements, (ii) Call for Paper announcements, (iii) Reading Materials \& Tutorials, (iv) Career Opportunities, (v) Talks \& Seminars, and (vi) Others. The Tweet Explorer feature utilizes these tweet categories. The detailed description of each topic is presented in Section~\ref{subsec:TweetExp}. 
    We experiment by fine-tuning a BERT-base\cite{Devlin2019BERTPO} classifier and twitter-roberta-base\citep{barbieri2020tweeteval} to predict the tweet topics. The BERT-base model\footnote{We also experimented with a zero-shot classifier but it underperformed the BERT-base classifier.} obtains the best test accuracy of 75\% on a small manually annotated dataset\footnote{each tweet was annotated by two ML/NLP students and inter annotator agreement computed using Cohen's $\kappa$=0.68}.
    \item \emph{Conference Page Builder}: The Conference Page Builder classifies a tweet either as discussing an ongoing conference or other topics. The module builds specific conference pages using such tweets. 
    \item \emph{CFP Timeline Builder}: \label{subsec:CFPTB} The module processes `Call for Papers' tweets identified by the Tweet Classifier module. It extracts the conference (and workshop) name by regex-based keyword matching against a pre-compiled list of venues. The submission date are extracted from the tweets by labeling dates using the Spacy\footnote{\url{https://spacy.io/}} library. The tweets are arranged in a timeline sorted by the submission deadline.
    \item \emph{Paper Tweet Linker}: The Paper Tweet Linker module maps specific tweets to research papers using regex matching of the paper title and paper URL. The Paper Tweet Visualizer uses these mappings to embed the tweets on the research paper page on NLPExplorer.
\end{enumerate}
The pipeline then stores the tweets in the database after processing by the above modules. We schedule our system to automatically curate the Twitter data daily and increase it to an hourly frequency during ongoing conferences.

\section{\sysname{} Features}
\label{sec:features}
\subsection{Tweet Explorer}
\label{subsec:TweetExp}
\begin{figure}[!htb]
    \centering
    \includegraphics[width=1.0\columnwidth]{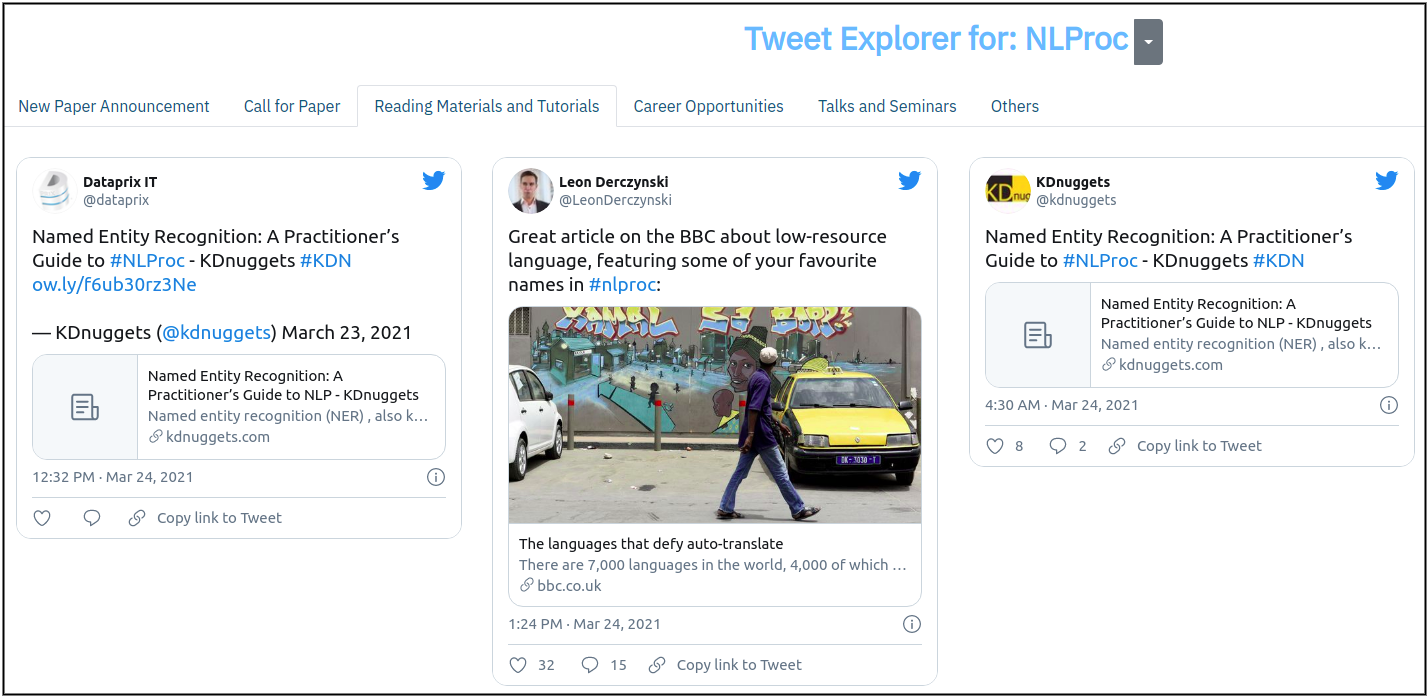}
    \caption{Tweet Explorer feature of \sysname{} which facilitates browsing tweets by six different topics.}
    \label{fig:TweetExp}
\end{figure}

We present a Tweet Explorer dashboard that allows a user to browse tweets from specific topics such as:
\begin{enumerate}[noitemsep,nolistsep]
    \item \emph{New paper announcements:} This topic organizes tweets about recent papers, which often involve the summary or a short introduction of the research paper. These twitter threads facilitate other researchers to communicate informally with the paper authors. These also contain interesting discussions by the community on the insights, merits, and critiques of the research paper, and post questions about the work. The authors' short introductions offer an informal account of the paper compared to the paper alert services that usually present the title and the abstract of the research paper.
    \item \emph{Call for Papers (CFPs) by various conferences and workshops:} Users can view the announcements for call for papers and submission deadlines by various workshops and conferences.
    \item \emph{Reading Materials \& Tutorials:} It lists various study material, such as lecture slides and videos, tutorials, online courses, and blog posts.
    \item \emph{NLP Career Opportunities:} Individuals frequently advertise opportunities for various positions such as interns, full-time, Ph.D., postdoctoral fellows, and research fellows on Twitter. 
    \item \emph{NLP Talks \& Seminars:} Various online NLP talks and seminars can be accessed using the NLP Talks \& Seminars filter on the Tweet Explorer dashboard.
    \item \emph{Others:} This category contains the NLP tweets which do not belong to any of the above topics.
\end{enumerate}
The Tweet Explorer feature allows users to specifically browse through tweets by topics and filter them based on their immediate interests. A snapshot of the same is presented in Figure~\ref{fig:TweetExp}. We present the distribution of tweets in the six categories from tweets curated by the system in the last 1,300 days in Table~\ref{tab:topic_dist}. 

\begin{table}[!htb]
\centering
\small{\begin{tabular}{|c|c|}
\hline
\textbf{Topic}                 & \textbf{Tweet Count} \\ \hline
New Paper Announcements        &    6,337                  \\ \hline
Call for Papers                &    972                    \\ \hline
Reading Materials \& Tutorials &    1,400                \\ \hline
NLP Career Opportunities       &    681                    \\ \hline
NLP Talks \& Seminars          &    2,382                \\ \hline
Others                         &    7,623             \\ \hline
\textbf{Total Tweets}          & \textbf{19,395}      \\ \hline
\end{tabular}}
\caption{Distribution of tweets (curated since October 2017) into various topics.}
\label{tab:topic_dist}
\end{table}

\begin{table*}[!ht]
\centering
\newcolumntype{A}{>{\centering\arraybackslash}m{0.13\linewidth}}
\newcolumntype{M}{>{\centering\arraybackslash}m{0.32\linewidth}}
\newcolumntype{X}{>{\centering\arraybackslash}m{0.34\linewidth}}
\renewcommand{\tabularxcolumn}[1]{m{#1}} % Align multi-line cells vertically! :O
\small{
\renewcommand{\arraystretch}{1.3} % for the vertical padding
\begin{tabularx}{\linewidth}{|A A M X|}
\Xhline{0.1em}
\textbf{Top Hashtags} & \textbf{Top Mentions} & \textbf{Top URLs} & \textbf{Top Papers Discussed} \\
\Xhline{0.1em}
\hline
\#acl2020nlp & @aclmeeting & \href{https://virtual.acl2020.org/socials.html}{virtual.acl2020.org/socials.html} & Beyond Accuracy: Behavioral Testing of NLP models with CheckList \\
\hline
\#acl2020en & @emilymbender & \href{virtual.acl2020.org/plenary\_session\_keynote\_kathy\_mckeown.html}{virtual.acl2020.org/plenary\_ session\_keynote\_kathy\_mckeown.html} & Photon: A Robust Cross-Domain Text-to-SQL System \\
\hline
\#nlproc & @akoller & \href{https://virtual.acl2020.org/paper_main.701.html}{virtual.acl2020.org/paper\_main.701.html} & Climbing towards NLU: On Meaning, Form, and Understanding in the Age of Data \\
\hline
\#acl2020zht & @winlpworkshop & \href{http://virtual.acl2020.org/workshop_W1.html}{virtual.acl2020.org/workshop\_W1.html} & Language Models as an Alternative Evaluator of Word Order Hypotheses: A Case Study in Japanese  \\
\hline
\#acl2020hi & @xandaschofield & \href{https://www.aclweb.org/anthology/2020.acl-main.442/}{www.aclweb.org/anthology/ 2020.acl-main.442/} & Don't Stop Pretraining: Adapt Language Models to Domains and Tasks \\
\hline
\#mt & @gneubig & \href{http://virtual.acl2020.org/workshop_W10.html}{virtual.acl2020.org/workshop\_W10.html} & The State and Fate of Linguistic Diversity and Inclusion in the NLP World \\
\hline
% \#winlp2020 & @sebgehr & \href{http://virtual.acl2020.org/paper_tacl.1915.html}{virtual.acl2020.org/paper\_tacl.1915.html} & Tangled up in BLEU: Reevaluating the Evaluation of Automatic Machine Translation Evaluation Metrics \\
% \hline
% \#acl2020ar & @dirk\_hovy & \href{https://alvr-workshop.github.io/}{alvr-workshop.github.io/} & Zero-Shot Transfer Learning with Synthesized Data for Multi-Domain Dialogue State Tracking \\
% \hline
% \#acl2020id & @msftresearch & \href{https://slideslive.com/38931667/t6-commonsense-reasoning-for-natural-language-processing}{slideslive.com/38931667} & Language (Technology) is Power: A Critical Survey of ``Bias'' in NLP \\
% \hline
% \#acl2020ko & @boknilev & \href{https://www.aclweb.org/anthology/2020.acl-main.463/}{www.aclweb.org/anthology/2020.acl-main.463/} & @Viraj, the current list only had Top9. Could you please add one more? \\
% \hline
\end{tabularx}}
\caption{ACL 2020 Twitter Coverage: Top discussed papers, mentions and URLs and popular hashtags.}
\label{tab:top_insights}
\end{table*}

\begin{table*}[!ht]
\centering
\renewcommand{\arraystretch}{1.3}% for the vertical padding!
\small{
\begin{tabular}{|c|c|c|c|c|}
\hline
\textbf{Tweet Counter} & \textbf{Likes Counter} & \textbf{Retweet Counter} & \textbf{Unique Mentions} & \textbf{Unique Paper Mentions}\\ \hline
5,343 & 58,160 & 11,440 & 907 & 251\\ \hline
\end{tabular}}
\caption{ACL 2020 Twitter Statistics of various activities.}
\label{tab:acl2020_stats}
\end{table*}

\subsection{Conference Visualizer -- Near real-time view for conferences}
\begin{figure*}[!htb]
\begin{center}
\includegraphics[width=.45\textwidth]{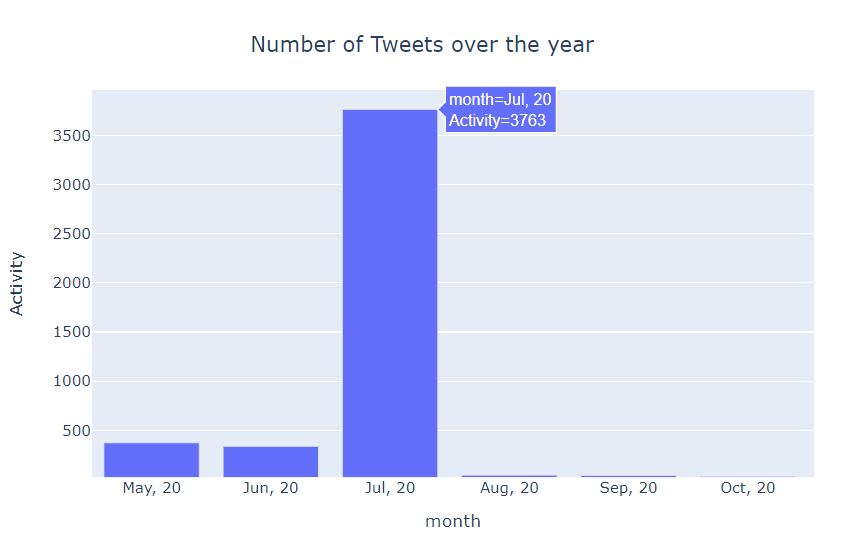}
\includegraphics[width=.45\textwidth]{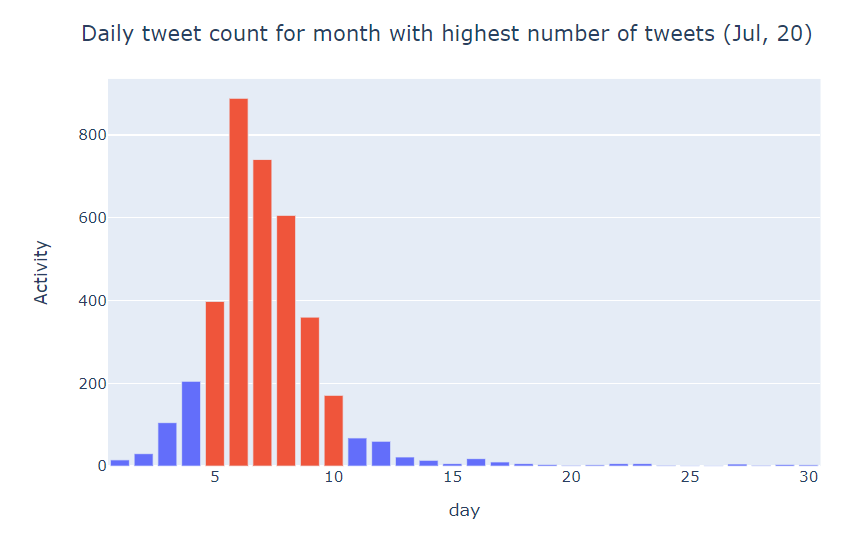}
\includegraphics[width=.45\textwidth]{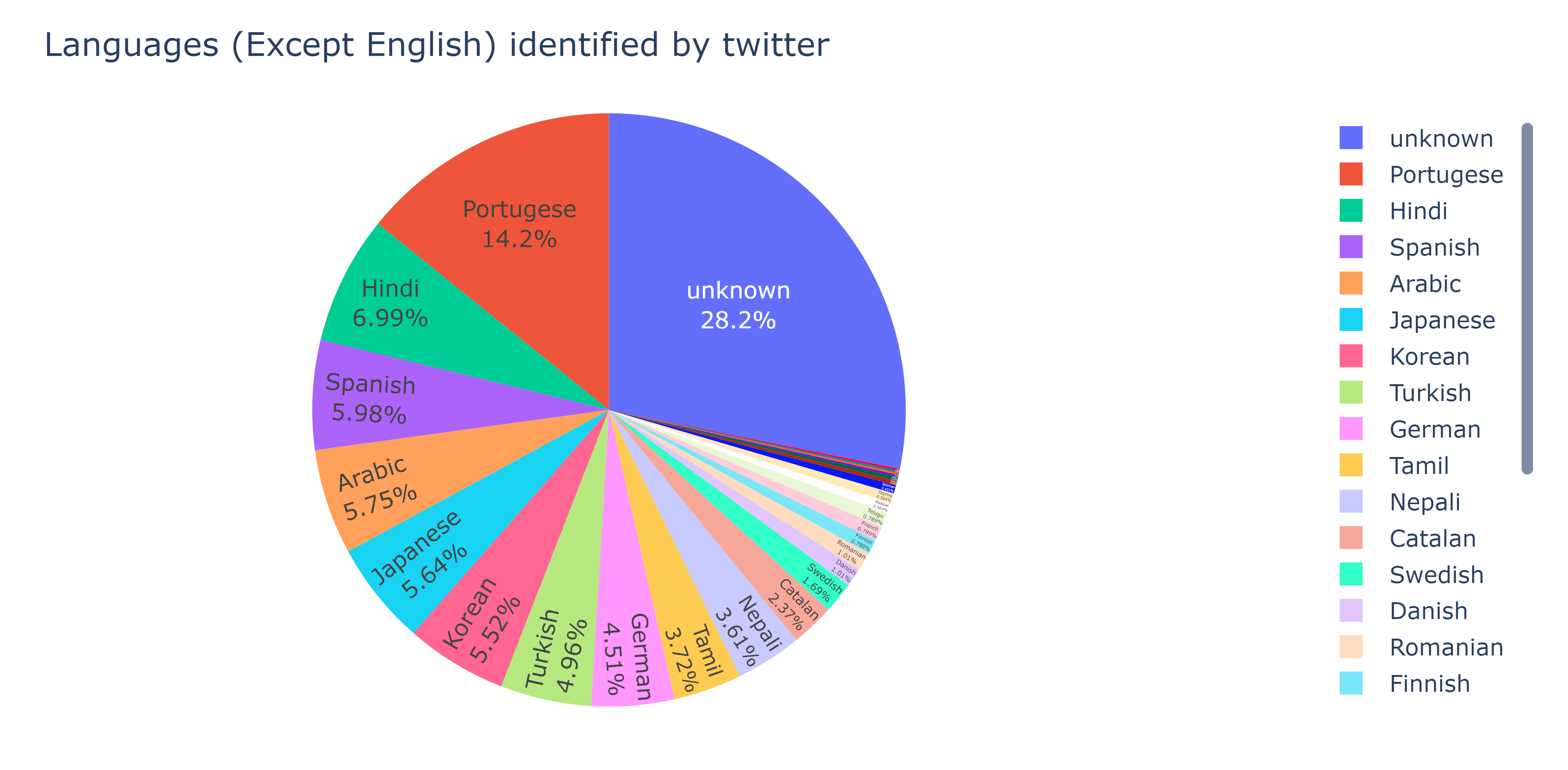}
\includegraphics[width=.45\textwidth]{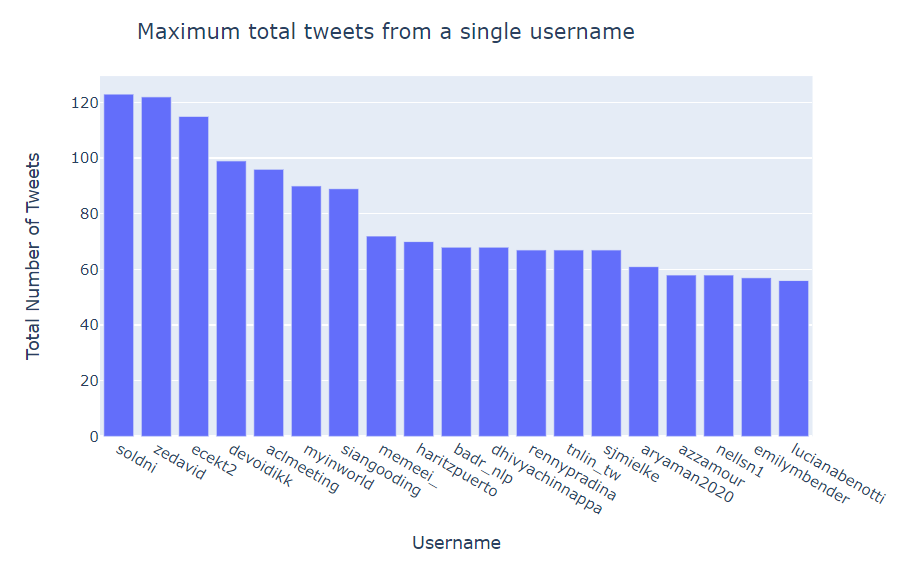}
\end{center}
\caption{Conference Visualizer: ACL2020 Statistics. (a) Distribution of tweets across different months. (b) Daily distribution of ACL2020 tweets in July 2020. (c) Distribution of tweet languages except English. (d) Twitter users with highest ACL2020 tweets. }
\label{fig:acl_vis}
\end{figure*}

\sysname{} supports real-time statistics for multiple top conferences and the popular \#NLProc hashtag. The information is updated hourly for live events and weekly for past events. Some of the statistics presented are top mentions, top hashtags, top linked URLs, and top discussed papers in tweets. We present the most popular hashtags, mentions, URLs, and highly discussed papers for ACL2020 in Table~\ref{tab:top_insights}. A summary of Twitter activity from the Conference Visualizer page for ACL 2020 is presented in Table~\ref{tab:acl2020_stats}. Apart from Twitter discussions about a conference in a specific month, we also show insights from the conferences across the year. The insights from ACL conference over time is presented in Figure~\ref{fig:acl_vis}.
We also present other conference-specific statistics such as the number of tweets per month, daily distribution of tweets in the conference month, most active users tweeting about the conference, and a distribution of the tweet languages other than English.

\subsection{Popular Paper Visualizer}
\begin{figure*}[!htb]
    \centering
    \includegraphics[width=1.7\columnwidth]{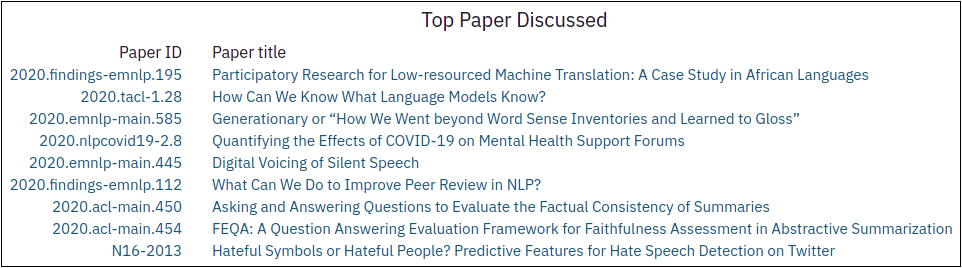}
    \caption{Popular papers identified \sysname{} based on twitter activity.}
    \label{fig:PopPaperVis}
\end{figure*}
We showcase widely discussed papers on Twitter in the Popular Paper Visualizer dashboard. It presents the titles and provides direct links to the full-text of the top discussed papers for quick reference. The system extracts tweets mentioning research papers and assigns a popularity score to each paper based on the count of tweets that mention it, and the likes, retweets, and replies on the paper tweets. We present a snapshot of few popular papers identified by our platform in Figure~\ref{fig:PopPaperVis}. It also presents the most active users tweeting about \#NLProc on Twitter. Popular Paper Visualizer dashboard also supports exploration of most liked and retweeted \#NLProc tweets of all times and in the last month.

\subsection{CFP Timeline}
\begin{figure}[!h]
    \centering
    \includegraphics[width=\columnwidth]{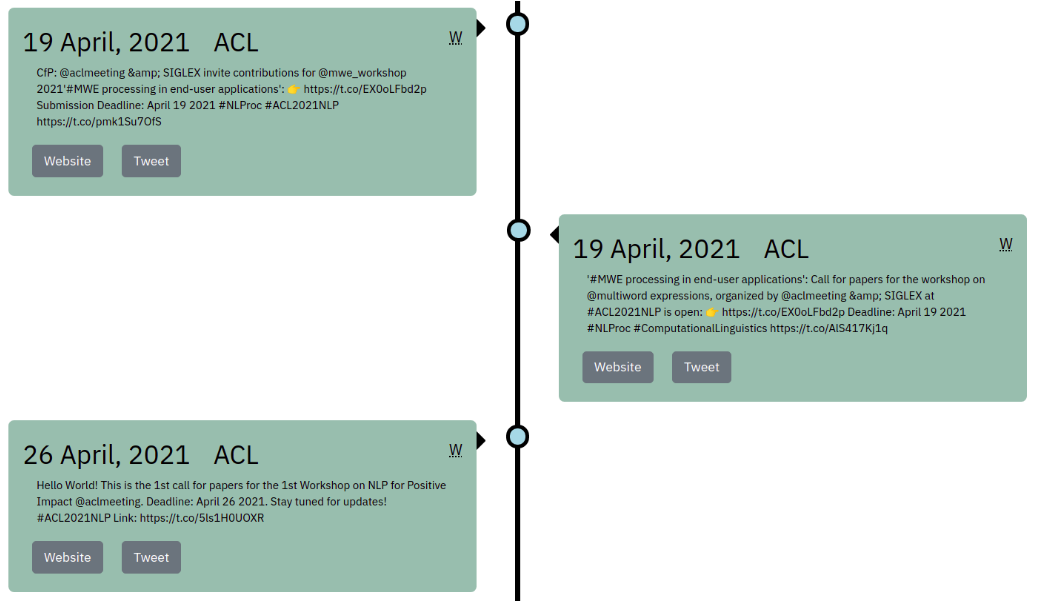}
    \caption{CFP Timeline built from tweets. `W' on top-right denotes Workshop.}
    \label{fig:cfp}
\end{figure}

\sysname{} presents a timeline of the upcoming submission deadlines. The timeline is created by identifying `Call for Papers' tweets using keyword based filtering of tweets and also lists the conference/workshop website. The details are described in the CFP Timeline Builder module \ref{subsec:CFPTB}. We present a snapshot of the timeline in Figure~\ref{fig:cfp}.

\subsection{Paper Tweet Visualizer}
\begin{figure}[!h]
    \centering
    \includegraphics[width=0.8\columnwidth]{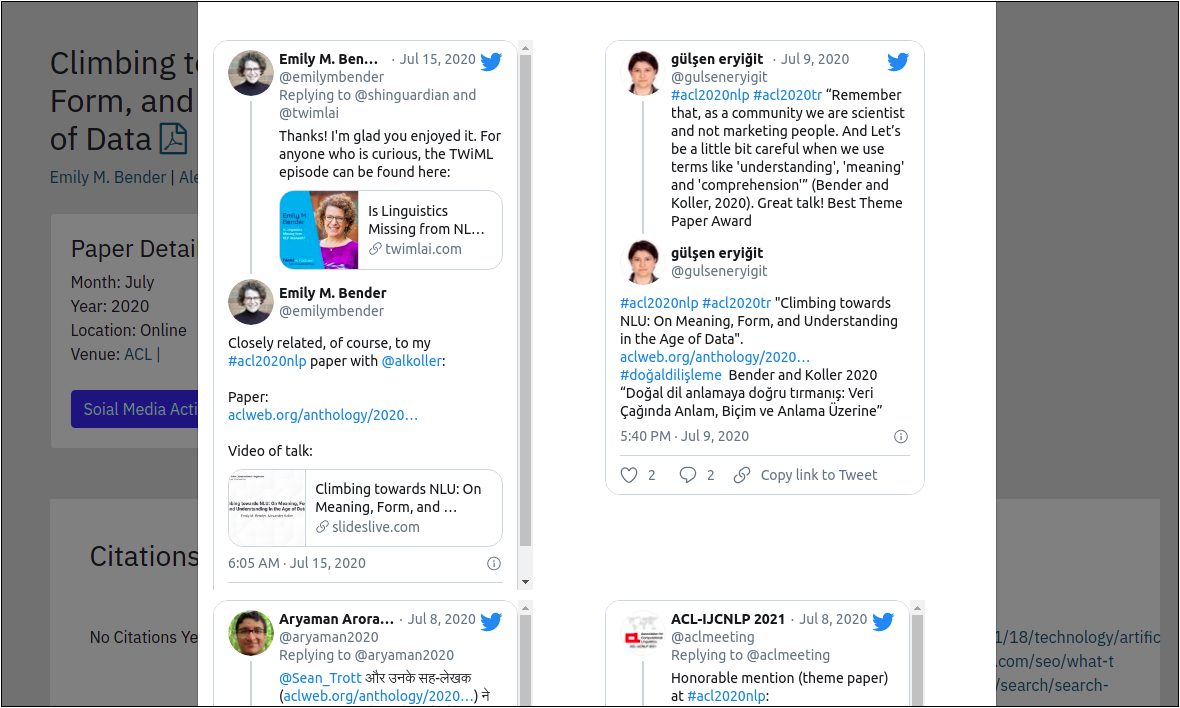}
    \caption{Paper Tweet Visualizer curates tweets and metadata of a research paper on a joint portal. The image background is a `Paper' page from NLPExplorer which lists paper metadata, citing papers, field-of-study tags, and similar papers alongwith the associated tweets.}
    \label{fig:PaperTweetVis}
\end{figure}

NLPExplorer supports a research paper search interface and builds research paper pages which showcase standard paper related statistics such as the publication year and venue, author information, citations, citation distribution over the years and the link to the corresponding PDF article. Additionally, it also provides interesting insights like similar papers, topical distribution and mentioned URLs. We map research paper discussion tweets on Twitter to the NLPExplorer paper page. This feature allows users to browse through discussions about the paper along with the metadata of the paper. We present a snapshot of the feature in Figure~\ref{fig:PaperTweetVis}.

\subsection{Popular Last Week}
Lastly, we present popular tweets in the NLP community on Twitter (also referred as NLP Twitter). This feature allows researchers to catch up with the recent NLP-related Twitter discussions in a single dashboard without searching for them specifically in the Twitter feed. 

\section{Related Works}
\label{sec:relwork}
\citet{bird2008acl} curated the \emph{ACL Anthology Reference Corpus (ACL ARC)} of research papers in NLP and CL. \citet{radev2009acl, radev2013acl} constructed the ACL Anthology Network (AAN) by manual annotation of the references to complete the citation network and analysed the network to present central papers, authors and other network statistics~\citep{radev2016bibliometric}. Works by \citet{schafer-etal-2011-acl} and \citet{parmar2020nlpexplorer} provide a comprehensive search interface to browse through the NLP based on parameters such as author, full text, year of publication, title, and the field of study. 
% Few works have built interfaces to visualize NLP research papers. 
\citet{mohammad-2020-nlp} built the NLPScholar platform which consists of interactive dashboards that present various aspects of NLP research papers. The platform uses ACL Anthology and Google Scholar as the information source. 

Few works have analysed Twitter data to predict scholarly impact. \citet{shuai2012scientific} 
% analyzed 4606 scientific articles submitted to arXiv.org and 
report a statistical correlation between high volume of Twitter mentions and arXiv downloads and early citations (i.e., citations occurring less than seven months after the publication of a preprint). However, they also point out that Twitter mentions cannot be directly concluded to be causative of higher levels of download and early citations.
Several other works such as~\citet{eysenbach2011can}, \citet{thelwall2013altmetrics}, and \citet{haustein2014tweeting} have tried to analyze whether tweets correlate with citations.

However, to the best of our knowledge, no prior work has tried to curate NLP discussions data from Twitter in an attempt to organize it and link it to research papers via a search engine or a visualization portal.

\section{Future Scope and Extensions}
\label{sec:futurescope}
Currently, the system is implemented only for NLP papers present in the ACL Anthology. The system could be extended to papers from NeurIPS, ICLR, and CVPR as the data for these conferences is available publicly. The system is versatile and can be easily extended to other domains. \sysname{} provides basic visualization graphs over Twitter activity. Over time, these discussions could be used to build a timeline of evolution of research in various domains of NLP based on the Twitter activity of researchers. Tweets by popular users attain likes and retweets at a higher rate in comparison to new users (or users with less followers) of the community.~\sysname{} currently only presents popular tweets based on retweets and likes count which can bias the conversations, understanding and presentation of ideas by emphasising the tweets of a small set of popular users. Future work includes identifying novel alternative ideas and perspectives by adjusting user popularity to create an inclusive space for the community.

\bibliographystyle{acl_natbib}
\bibliography{anthology,acl2021}

%\appendix

\end{document}